\documentclass[11pt,a4paper]{article}

\usepackage[T1]{fontenc}
\usepackage[utf8]{inputenc}
\usepackage{mathpazo}
\usepackage{microtype}

\usepackage[a4paper,margin=2.5cm]{geometry}
\usepackage{amsmath,amssymb}

\usepackage{graphicx}
\usepackage{booktabs}
\usepackage{array}
\usepackage{tabularx}
\usepackage{ragged2e}
\usepackage{float}
\usepackage{caption}
\usepackage{placeins}
\usepackage{fancyhdr}
\usepackage[inline]{enumitem}
\setlist[itemize]{
  leftmargin=1.6em,
  noitemsep,
  topsep=3pt,
  parsep=0pt
}
\setlist[itemize,2]{leftmargin=1.8em}
\setlist[itemize,3]{leftmargin=2.0em}
\setlist[description]{
  leftmargin=2.2cm,
  labelsep=0.6em,
  font=\normalfont\bfseries,
  style=nextline
}

\usepackage[numbers,sort&compress]{natbib}
\usepackage{xurl}

\usepackage{xcolor}
\definecolor{linkblue}{RGB}{0,70,120}

\usepackage{hyperref}
\hypersetup{
  colorlinks=true,
  linkcolor=linkblue,
  citecolor=linkblue,
  urlcolor=linkblue,
  pdftitle={
    TRACTA: Benchmarking Temporal Reasoning over Semantic Trajectories
  },
  pdfauthor={
    Michael Romei de Socio,
    Gian Luca Pozzato,
    Alessio Merlo
  }
}

\usepackage{authblk}
\usepackage{orcidlink}

\title{
  TRACTA: Benchmarking Temporal Reasoning\\
  over Semantic Trajectories
}

\author[1,2]{
  Michael Romei de Socio\,
  \orcidlink{0009-0008-3949-0437}
  \thanks{
    Email addresses:
    \href{mailto:michael.romeidesocio@unito.it}
    {michael.romeidesocio@unito.it};
    \href{mailto:michael.romeidesocio@unicasd.it}
    {michael.romeidesocio@unicasd.it}
  }
}

\author[1]{
  Gian Luca Pozzato\,
  \orcidlink{0000-0002-3952-4624}
  \thanks{
    Principal corresponding author:
    \href{mailto:gianluca.pozzato@unito.it}
    {gianluca.pozzato@unito.it}
  }
}

\author[2]{
  Alessio Merlo\,
  \orcidlink{0000-0002-2272-2376}
  \thanks{
    Corresponding author:
    \href{mailto:alessio.merlo@unicasd.it}
    {alessio.merlo@unicasd.it}
  }
}

\affil[1]{
  Department of Computer Science,
  University of Turin,
  Corso Svizzera 185,
  10149 Turin, Italy
}

\affil[2]{
  CASD -- School of Advanced Defense Studies,
  Piazza della Rovere 83,
  00165 Rome, Italy
}

\date{\small Preprint -- September 2026}

\begin{document}

\maketitle

\begin{abstract}
High-complexity operational environments require methods that characterize temporally distributed patterns rather than classify isolated events. This paper introduces TRACTA (Temporal Reasoning and Capability-Trajectory Analysis), a knowledge-aligned synthetic benchmark for temporal structural reasoning, instantiated through Multi-Domain Operations (MDO)-like scenarios. TRACTA defines offline structural annotations over contextual direct-impact and accumulated capability trajectories and evaluates three tasks: \texttt{early\_\allowbreak warning}, \texttt{pattern\_\allowbreak detection}, and \texttt{run\_\allowbreak classification}. The frozen comparison includes raw-event neural references, a contract-lite rule comparator, and a recurrent semantic-input reference.

The semantic-input recurrent reference has the highest aggregate macro-F1 point estimates, with the largest margins on the two temporal tasks, while raw-event references remain predictive and lead in four individual early-warning target--lead settings. Component-zeroing diagnostics show that both semantic trajectory blocks contain useful signal within the evaluated recurrent configuration. Run-local aliasing removes stable cross-run target and location identities from the primary raw input, although executed diagnostics retain shallow predictivity. These results are configuration-level: semantic inputs are aligned with the benchmark's target-generation space, and the evaluated systems also differ in architecture, training, and available information. TRACTA therefore provides a reproducible testbed for examining knowledge-aligned temporal prediction, not evidence of a causal representation advantage, statistically resolved superiority, or operational readiness.
\end{abstract}


\bigskip

\section{Introduction}
\label{sec:introduction}

High-complexity operational environments increasingly involve heterogeneous events interacting across multiple domains, infrastructures, and decision layers. In such settings, decision-support systems are required not only to process individual observations but also to assess whether temporally distributed evidence reflects the emergence of structurally meaningful patterns.

This challenge is particularly evident in Multi-Domain Operations (MDO), where cyber, physical, and informational effects may propagate across interdependent capabilities. Individual events---such as disruptions, delays, or localized actions---are often weak or ambiguous when considered in isolation. However, their combined temporal and contextual evolution may indicate the degradation of higher-level system functions. Addressing this problem therefore requires reasoning over temporal accumulation, contextual impact, and interactions among capabilities.

Raw event-level neural models provide a natural baseline, as they can learn temporal dependencies directly from sequential data. At the same time, such models may be sensitive to superficial regularities, especially in controlled or synthetic datasets. Semantic and rule-based approaches offer a complementary perspective by encoding domain knowledge and enabling interpretable reasoning over capability dependencies, but they may have limited ability to capture noisy, delayed, or weakly distributed temporal signals.

These considerations motivate a neuro-symbolic formulation that combines structured semantic representations with data-driven temporal learning. In this setting, the semantic component transforms event-level observations into contextualized capability and impact trajectories, while the learning component models their temporal evolution.
This organization provides an inspectable interface between event data and temporal learning, while allowing accumulated and contextualized capability states to be modeled over time.

To study this formulation under controlled conditions, we introduce \emph{TRACTA} (\emph{Temporal Reasoning and Capability-Trajectory Analysis}), a synthetic benchmark for temporal structural reasoning in high-complexity event-driven systems, instantiated through MDO-like scenarios.
TRACTA is deliberately knowledge-aligned: its semantic reference layer contributes both to the structured input trajectories and to the procedure used to define the prediction targets. The benchmark therefore evaluates a controlled configuration-level comparison rather than isolating a causal effect of semantic representation.

The benchmark evaluates three tasks:
\texttt{early\_\allowbreak warning}, which predicts a future structural-label state from a preceding observation window;
\texttt{pattern\_\allowbreak detection}, which predicts the offline structural annotation indexed at the reference time;
and \texttt{run\_\allowbreak classification}, which provides a coarse full-run assessment.
The temporal labels are generated offline from complete synthetic trajectories, including retrospective persistence and future recovery conditions, while the primary predictive inputs remain restricted to their task-specific observation windows.

The experimental setting compares raw event-level neural references, including a Transformer-based configuration, a contract-lite semantic rule comparator, and a recurrent semantic-input configuration operating on semantically grounded trajectories.
The objective is to compare these evaluated configurations under shared datasets, partitions, task definitions, and metrics. Because their architectures, training settings, available information, and representations do not fully match, the experiments cannot attribute performance differences solely to semantic representation.

TRACTA and its benchmark methodology constitute the main contribution; the evaluated systems serve as reference configurations rather than as a proposed novel recurrent architecture.

\begin{figure*}[t]
    \centering
    \includegraphics[width=\textwidth]{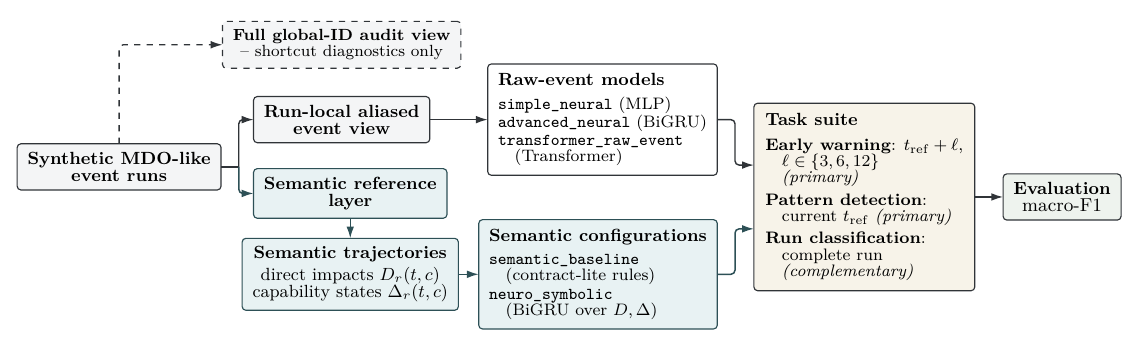}
    \caption{Representation-aware TRACTA comparison. The full global-ID
    view is retained only for shortcut diagnostics. Raw-event models
    operate on the run-local aliased event view, whereas the semantic
    rule comparator and semantic-input recurrent reference use contextual
    direct-impact and accumulated capability trajectories generated by
    the semantic reference layer. All configurations are evaluated on
    the same task suite, although their information content, architectures,
    and training settings are not fully matched.}
    \label{fig:representation-aware-comparison}
\end{figure*}

Figure~\ref{fig:representation-aware-comparison} summarizes the input views, reference configurations, common task suite, and diagnostic-only role of the full global-identifier view. The offline construction of the temporal targets is detailed in Figure~\ref{fig:offline-target-construction}.

The evaluated semantic-input recurrent configuration achieves the highest aggregate point estimates, with the largest margins on the two temporal tasks. Raw event-level learning nevertheless remains informative, and comparator configurations lead in four individual early-warning target--lead settings.

These findings characterize the evaluated configurations within a fixed synthetic benchmark. They do not establish statistically resolved superiority, a causal representation effect, generalization to unseen scenario families, or operational readiness.

Overall, TRACTA provides a reproducible testbed for studying how event-level inputs, knowledge-aligned state trajectories, rule-based reasoning, and temporal learning interact under explicitly documented benchmark conditions.

\section{Related Work}
\label{sec:related_work}

\subsection{Neuro-Symbolic AI and Knowledge-Guided Learning}

Neuro-symbolic artificial intelligence encompasses approaches that combine neural learning with structured knowledge, including logical constraints, ontologies, and knowledge bases~\cite{hitzler_neuro_symbolic_2022,sarker_neuro_symbolic_2022,garcez_neurosymbolic_2023}. Rather than defining a single architecture, this literature includes systems that use symbolic knowledge to structure inputs, constrain learning, support inference, or improve interpretability.

Representative frameworks illustrate different integration approaches. DeepProbLog combines neural predicates with probabilistic logic programming~\cite{manhaeve_deepproblog_2021}, while Logic Tensor Networks encode logical constraints within differentiable models~\cite{badreddine_ltn_2022}. These approaches integrate learning and reasoning more tightly than TRACTA, whose semantic reference layer is fixed rather than jointly optimized with the predictive models.

Knowledge-guided and theory-guided machine learning similarly incorporate domain knowledge through features, constraints, architectures, loss functions, or intermediate representations~\cite{dash_domain_knowledge_2022,karpatne_theory_guided_2017,willard_scientific_knowledge_2022}.
TRACTA follows this orientation as a knowledge-aligned benchmark: a fixed semantic reference layer maps events into contextual direct-impact and accumulated capability trajectories, which are then used by rule-based and recurrent reference configurations. Because the same modeled capability space contributes to target construction, the study compares complete configurations rather than isolating a causal effect of semantic representation.

\subsection{Temporal Pattern Detection and Early Warning}

TRACTA is related to time-series classification and sequence learning, where convolutional, recurrent, and hybrid architectures learn temporal representations from sequential observations~\cite{fawaz_tsc_review_2019}. Architectures such as InceptionTime also emphasize the need for credible temporal references when evaluating alternative representations or model designs~\cite{fawaz_inceptiontime_2020}.

The \texttt{early\_warning} task is connected to early time-series classification, in which predictions are made before the complete sequence has been observed and performance reflects a trade-off between earliness and accuracy~\cite{xing_early_classification_2011,achenchabe_early_classification_2021}. Early warning has also been studied with heterogeneous sensor streams, including earthquake monitoring~\cite{fauvel_earthquake_early_warning_2020}.
TRACTA differs from standard early classification because it predicts future states of offline structural annotations defined through capability degradation, event-context filters, retrospective persistence, and recovery conditions. The task does not exclusively predict the onset of a new episode, nor does it evaluate prospective operational alert utility.

Attention-based sequence models provide an additional architectural reference. Transformers introduced self-attention as an alternative to recurrent sequence modeling and have subsequently been applied to multivariate time-series representation learning~\cite{vaswani_attention_2017,zerveas_time_series_transformer_2021}.
TRACTA therefore includes \texttt{transformer\_raw\_event} as an additional attention-based raw-event reference. Its inclusion broadens the evaluated architecture classes, but does not exhaust competitive event models or separate representation effects from architecture, optimization, and information-access differences.

\subsection{Synthetic Benchmarks, Shortcut Learning, and Evaluation}

Synthetic data and simulation-based benchmarks are useful when the target phenomenon is difficult to observe, annotate, or manipulate under controlled conditions~\cite{nikolenko_synthetic_data_2021}. Their interpretation nevertheless depends on construction assumptions, documentation, and intended use~\cite{paullada_dataset_development_2021,bender_data_statements_2018}. TRACTA is therefore positioned as a controlled methodological benchmark rather than as operational validation.

A central concern is shortcut learning, whereby models exploit predictive cues that do not correspond to the intended construct~\cite{geirhos_shortcut_2020}. Dataset bias and leakage can similarly inflate benchmark performance or undermine reproducibility~\cite{torralba_dataset_bias_2011,kapoor_leakage_2023}.
TRACTA removes stable cross-run target and location identities from the primary raw input through run-local aliasing, while retaining the full global-identifier view for diagnostics. This intervention removes a specific identity channel by construction; it does not establish that the benchmark is free of shallow predictivity, nor that the main models rely on the mechanisms the benchmark intends.

Robust benchmark reporting also requires attention to experimental variance, uncertainty, and reproducibility~\cite{diciccio_bootstrap_1996,ovadia_uncertainty_shift_2019,bouthillier_variance_benchmarks_2021,pineau_reproducibility_2021}.
TRACTA addresses these concerns through run-level partitioning, multiple training seeds, diagnostic comparisons, component-zeroing conditions, grouped-bootstrap uncertainty summaries, and frozen reproducibility artifacts. These measures improve traceability but do not establish robustness to alternative splits, unseen scenario families, repeated benchmark generation, or statistically resolved differences between configurations.

\subsection{High-Complexity Operational Environments}

TRACTA's operational motivation comes from decision-support settings in which heterogeneous events interact over time and affect interdependent system components. Situation-awareness research frames such settings through perception, comprehension, and projection of evolving states~\cite{endsley_situation_awareness_1995}, while complex-systems research emphasizes that distributed interactions can generate dynamics not reducible to isolated events or actors~\cite{bak_coleman_collective_behavior_2021}. In the defense domain, Multi-Domain Operations describe coordinated activity across physical and non-physical domains~\cite{nato_ajp01_2022}.

This literature motivates MDO-like scenarios as one controlled instantiation of temporal structural reasoning in high-complexity systems. TRACTA is neither a doctrinal MDO model nor an operational decision-support system. It contributes a synthetic, knowledge-aligned testbed that places raw-event, contract-lite rule, and semantic-input recurrent references under shared tasks and partitions, while explicitly documenting configuration differences and validity boundaries.

\section{Problem Framing and Benchmark Rationale}
\label{sec:problem-framing}

This section defines the decision-support problem TRACTA addresses and motivates the benchmark design. The central construct is \emph{structural convergence}: the emergence of temporally and contextually meaningful degradation patterns over capability trajectories, rather than separability over isolated event tokens or superficial event statistics.

\subsection{High-Complexity Operational Trajectories}
\label{subsec:high-complexity-operational-trajectories}

TRACTA is motivated by settings in which relevant signals are distributed across time, targets, locations, actors, and capability dependencies. This framing aligns with situation-awareness research and the treatment of Multi-Domain Operations (MDO) as a joint, multi-domain operating context~\cite{endsley_situation_awareness_1995,nato_ajp01_2022}. In such environments, cyber disruptions, electromagnetic interference, logistics delays, kinetic actions, information effects, infrastructure degradation, or changes in coalition support may be meaningful locally but insufficient on their own.

The analytical problem is whether heterogeneous observations jointly indicate a broader degradation pattern. A localized event may become significant only when combined with events affecting related capabilities, occurring within reinforcing temporal windows, or propagating through shared dependencies. TRACTA therefore uses MDO-like trajectories as a controlled synthetic instance of a broader problem: predicting structurally defined states in temporally evolving operational systems.

\subsection{Structural Convergence as the Target Construct}
\label{subsec:structural-convergence}

A trajectory is structurally convergent when heterogeneous events, distributed across time and operational contexts, jointly produce capability-level degradation patterns satisfying the benchmark label conditions. The target is therefore not a specific event type, subtype, or entity token, but a pattern defined over the modeled operational state.

Let $r$ denote a run with event sequence
\begin{equation}
E_r=(e_{r,1},\ldots,e_{r,n_r}),
\label{eq:run-sequence}
\end{equation}
where each event includes a timestamp, event type, subtype, target, location, actor, confidence, severity, and optional notes. The semantic reference layer maps $E_r$ into two trajectory families over a finite capability set $\mathcal{C}$ and time horizon $\mathcal{T}$:
\begin{equation}
D_r(t,c), \qquad \Delta_r(t,c),
\qquad t\in\mathcal{T},\ c\in\mathcal{C}.
\end{equation}
Here, $D_r(t,c)$ is the contextual direct-impact trajectory, aggregating resolved event-level impacts on capability $c$ at time $t$, while $\Delta_r(t,c)$ is the accumulated capability trajectory after direct impacts, propagation, persistence, decay, and clipping. Their construction is detailed in Appendix~\ref{app:pattern-formalization}.

Let $\mathcal{K}=\{1,2,3\}$ denote the pattern identifiers. For each $k\in\mathcal{K}$, the capability-state condition is
\begin{equation}
S_{r,k}(t)=\psi_k\bigl(\Delta_r(t,\cdot)\bigr),
\label{eq:capability-state-condition}
\end{equation}
where $\psi_k$ encodes the thresholds and logical combinations associated with pattern $k$. It is combined with the structural context filter
\begin{equation}
Q_{r,k}(t)=\chi_k(E_r,t),
\label{eq:structural-context-filter}
\end{equation}
where $\chi_k$ checks cross-domain, cross-target, location-cluster, and adverse-pressure evidence, together with the future recovery condition detailed in Appendix~\ref{app:structural-filter}.
The raw activation mask is
\begin{equation}
m_{r,k}(t)=S_{r,k}(t)\land Q_{r,k}(t).
\label{eq:raw-activation}
\end{equation}
The temporal pattern label applies a persistence operator:
\begin{equation}
p_{r,k}(t)=\operatorname{Consecutive}\bigl(m_{r,k},d_k\bigr)(t),
\qquad k\in\mathcal{K},
\label{eq:pattern-persistence}
\end{equation}
where $d_k$ is the minimum active duration required by the corresponding pattern contract.
The operator labels every position in a contiguous active segment whose total duration reaches $d_k$; early positions in a qualifying segment are therefore confirmed retrospectively.
The global temporal label is
\begin{equation}
z_r(t)=\mathbf{1}\left[\bigvee_{k\in\mathcal{K}}p_{r,k}(t)\right],
\label{eq:global-temporal-label}
\end{equation}
and the semantic run-level label is
\begin{equation}
y_r^{\mathrm{sem}}=\mathbf{1}\left[\exists t\in\mathcal{T}:z_r(t)=1\right].
\label{eq:semantic-run-label}
\end{equation}

This formalization defines the benchmark targets through capability degradation, structural context, and persistence. A run may contain disruptive events yet remain negative when these conditions are not jointly satisfied. These definitions specify the intended construct, but do not establish that a fitted model relies on the corresponding mechanism rather than on correlated statistics.

\subsection{Comparative Modeling Rationale}
\label{subsec:comparative-modeling-rationale}

Raw event-level learning is a necessary comparator because event logs contain useful temporal and contextual information. However, synthetic benchmarks may contain stable identifiers, recurring locations, temporal templates, or repeated entity combinations that act as unintended proxy labels~\cite{geirhos_shortcut_2020,kapoor_leakage_2023}.

Semantic and rule-based approaches encode relations among events, targets, locations, and capabilities, making the modeled propagation of local impacts explicit. This supports interpretability and aligns with neuro-symbolic and knowledge-guided learning~\cite{hitzler_neuro_symbolic_2022,willard_scientific_knowledge_2022}. Fixed rules may nevertheless be sensitive to calibration and may underuse distributed temporal evidence. The semantic configuration is therefore a contract-lite comparator, not a symbolic oracle.

The recurrent semantic-input configuration combines the fixed semantic reference layer with temporal learning over capability and contextual direct-impact trajectories.
TRACTA compares this configuration with raw-event neural references and the contract-lite rule comparator under shared tasks and partitions. Because their representations, available information, architectures, and training procedures are not fully matched, the comparison characterizes complete configurations and does not isolate a causal representation effect.

\subsection{Task Hierarchy and Interpretive Weight}
\label{subsec:task-hierarchy}

TRACTA includes \texttt{early\_\allowbreak warning}, \texttt{pattern\_\allowbreak detection}, and \texttt{run\_\allowbreak classification}. Let $H$ be the run horizon, $w$ the temporal window size, $t_{\mathrm{ref}}$ the reference time, and $\ell$ the forecasting lead. In TRACTA, $H=96$, $w=8$, and $\ell\in\{3,6,12\}$ for \texttt{early\_\allowbreak warning}. For temporal-window tasks, a model observes
\begin{equation}
X_r[t_{\mathrm{ref}}-w+1:t_{\mathrm{ref}}],
\end{equation}
where $X_r$ is the input representation of the evaluated configuration. Raw-event models use the run-local aliased event view; semantic configurations use capability and contextual direct-impact trajectories.

Let $\mathcal{I}=\{z\}\cup\mathcal{K}$ index the global and pattern-specific targets, with
\begin{equation}
y_{r,z}(t)=z_r(t), \qquad y_{r,k}(t)=p_{r,k}(t),\quad k\in\mathcal{K}.
\label{eq:task-target-mapping}
\end{equation}
For \texttt{early\_\allowbreak warning},
\begin{equation}
\hat{y}^{\mathrm{EW}}_{r,q}(t_{\mathrm{ref}},\ell)
=f^{\mathrm{EW}}_{\theta,q}\left(X_r[t_{\mathrm{ref}}-w+1:t_{\mathrm{ref}}]\right),
\end{equation}
with target
\begin{equation}
y^{\mathrm{EW}}_{r,q}(t_{\mathrm{ref}},\ell)
=y_{r,q}(t_{\mathrm{ref}}+\ell),
\qquad q\in\mathcal{I},\ \ell\in\{3,6,12\}.
\label{eq:early-warning-target}
\end{equation}
This task predicts a future offline label state from a preceding window. It does not exclusively predict episode onset, because the target pattern may already be active at $t_{\mathrm{ref}}$, and it does not evaluate prospective operational alert utility.

For \texttt{pattern\_\allowbreak detection},
\begin{equation}
\hat{y}^{\mathrm{PD}}_{r,q}(t_{\mathrm{ref}})
=f^{\mathrm{PD}}_{\theta,q}\left(X_r[t_{\mathrm{ref}}-w+1:t_{\mathrm{ref}}]\right),
\end{equation}
with target
\begin{equation}
y^{\mathrm{PD}}_{r,q}(t_{\mathrm{ref}})=y_{r,q}(t_{\mathrm{ref}}),
\qquad q\in\mathcal{I}.
\label{eq:pattern-detection-target}
\end{equation}
The two temporal tasks provide the primary configuration-level evidence for prediction of the benchmark's offline structural annotations.

The complementary \texttt{run\_\allowbreak classification} task operates on the complete run:
\begin{equation}
\hat{y}^{\mathrm{RC}}_r=f^{\mathrm{RC}}_{\theta}\left(X_r[0:H-1]\right),
\end{equation}
with target
\begin{equation}
y^{\mathrm{RC}}_r=\texttt{run\_\allowbreak label}(r).
\label{eq:run-classification-target}
\end{equation}
It evaluates coarse full-run separability and is treated as a complementary endpoint.

\subsection{Shortcut Risk and Benchmark Design}
\label{subsec:shortcut-leakage-threat}

Synthetic generation may introduce correlations that are easier to learn than the intended construct. In TRACTA, the main identified risk is reusing globally stable target and location identifiers. Let $g_T(e)$ and $g_L(e)$ denote these identifiers. The full raw audit view retains them, whereas the primary raw input applies the run-local map
\begin{equation}
a_r:\bigl(g_T(e),g_L(e)\bigr)\mapsto
\bigl(\widetilde{T}_r(e),\widetilde{L}_r(e)\bigr).
\end{equation}
The same entity therefore has no stable token across runs, while recurrence within a run is preserved.

Run-local aliasing removes the direct cross-run identity channel by construction, but does not make the benchmark leakage-free. Within-run recurrence, temporal organization, confidence, severity, paired-run structure, and other generator-dependent statistics may remain predictive. The executed diagnostics cover only selected views and tasks and do not identify the features used by the main neural models.

Accordingly, TRACTA is designed for a bounded comparison of raw-event, contract-lite rule, and semantic-input recurrent references on the same synthetic runs and task definitions. Its empirical analysis reports configuration-level point estimates and diagnostics, rather than testing a prespecified claim of neuro-symbolic superiority.

\section{Methodology: Semantic Reference Layer and Benchmark Construction}
\label{sec:methodology}

This section describes how we generate, represent, label, split, and audit TRACTA. Consistent with Section~\ref{sec:problem-framing}, the benchmark operationalizes structural convergence through synthetic runs, semantic trajectory construction, contract-based labeling, controlled dataset views, run-level partitioning, and shortcut diagnostics. Model configurations and evaluation procedures are described in Section~\ref{sec:experimental-setup}.

\subsection{Synthetic MDO-Like Runs}
\label{subsec:synthetic-rationale}

TRACTA uses synthetic MDO-like runs because operational event streams may be sensitive, incomplete, non-repeatable, or lack reliable labels for latent convergence patterns. Synthetic construction lets us document and control scenario logic, semantic resolution, target definitions, dataset views, and partitioning~\cite {nikolenko_synthetic_data_2021,paullada_dataset_development_2021}.

The purpose is not operational replication, but construction of a fixed testbed for evaluating temporal structural-reasoning configurations. The benchmark therefore characterizes performance under its explicit generation assumptions and does not establish validity for operational data.

The frozen corpus is produced by an acceptance-controlled paired generator. For each of five scenario families, a canonical event log provides event-type and subtype prototypes, while family specifications define phase budgets, entity pools, structural requirements, and class-specific compilation policies. The generator constructs 20 positive/negative pairs per family from a shared prototype-derived template and ordered timestamp schedule. A pair is retained only when both branches pass schema and structural validation and satisfy their prescribed pattern guardrails. The resulting corpus contains 200 runs, each with 96 events over a 96-step horizon. The semantic reference layer contains 20 capabilities, 26 dependency edges, 90 target instances, and 34 location instances.

\subsection{Semantic Reference Layer}
\label{subsec:semantic-reference-layer}

The semantic reference layer maps each event sequence $E_r$ into contextual direct-impact trajectories $D_r(t,c)$ and accumulated capability trajectories $\Delta_r(t,c)$ over $c\in\mathcal{C}$ and $t\in\mathcal{T}$. Raw-event and semantic views originate from the same event surface, but the semantic path treats targets and locations as variables that condition modeled capability effects.

For event $e$ and capability $c$, let $s(e)$ denote the event subtype. The contextual impact range is
\begin{equation}
\begin{aligned}
\widetilde{I}_{e,c}
=\operatorname{clip}_{[-1,1]}\Big(&I^{\mathrm{base}}_{s(e),c}
\,w_{\mathrm{target}}(e,c)
\,m_{\mathrm{crit}}(e)\\
&\,m_{\mathrm{loc}}(e,c)
\,m_{\mathrm{disloc}}(e)\Big),
\end{aligned}
\label{eq:contextual-impact-range}
\end{equation}
where $I^{\mathrm{base}}_{s(e),c}$ is the subtype-level impact range, $w_{\mathrm{target}}(e,c)$ is the target-capability weight, $m_{\mathrm{crit}}(e)$ is the target-criticality multiplier, $m_{\mathrm{loc}}(e,c)$ is the location-capability modifier, and $m_{\mathrm{disloc}}(e)$ accounts for target-location incompatibility. Appendix~\ref{app:pattern-formalization} provides the operational specification.

This is a deliberately knowledge-aligned design: modeled capability states contribute both to semantic inputs and to target construction. The benchmark evaluates configurations with access to these concepts; it does not establish their general superiority over arbitrary representations or domains.

\subsection{Trajectory Construction}
\label{subsec:trajectory-construction}

Each resolved event-capability impact is sampled from its contextual range using event severity and a small noise term, yielding $x_{e,c}$. The direct-impact trajectory is
\begin{equation}
D_r(t,c)=\sum_{\substack{e\in E_r:\\ \tau(e)=t}}x_{e,c},
\label{eq:direct-impact-trajectory}
\end{equation}
and corresponds to the stored \texttt{direct\_\allowbreak values}. The accumulated capability trajectory is initialized as
\begin{equation}
\Delta_r(0,c)=\operatorname{clip}_{[-1,1]}\bigl(D_r(0,c)\bigr),
\label{eq:capability-trajectory-initialization}
\end{equation}
and, for $t>0$, evolves according to
\begin{equation}
\begin{aligned}
\Delta_r(t,c)=\operatorname{clip}_{[-1,1]}\Big(&
(1-\lambda)\Delta_r(t-1,c)+D_r(t,c)\\
&+\sum_{p\in\mathrm{Pa}(c)}w_{p,c}\Delta_r(t-1,p)\Big).
\end{aligned}
\label{eq:capability-trajectory-recursion}
\end{equation}
Here, $\mathrm{Pa}(c)$ contains the parent capabilities of $c$, $w_{p,c}$ is the propagation weight, and $\lambda$ is the decay parameter.

The frozen benchmark uses $\lambda=0.20$ and 26 explicit edge weights ranging from 0.19 to 0.46; the nominal 0.30 fallback is inactive in the reported graph. These values are specification choices, and we do not evaluate robustness to alternative values.
Capability trajectories are clipped to $[-1,1]$ and stored as \texttt{delta\_\allowbreak values}.

\subsection{Contract-Based Pattern Labeling}
\label{subsec:contract-based-labeling}

\begin{figure*}[t]
    \centering
    \includegraphics[width=\textwidth]{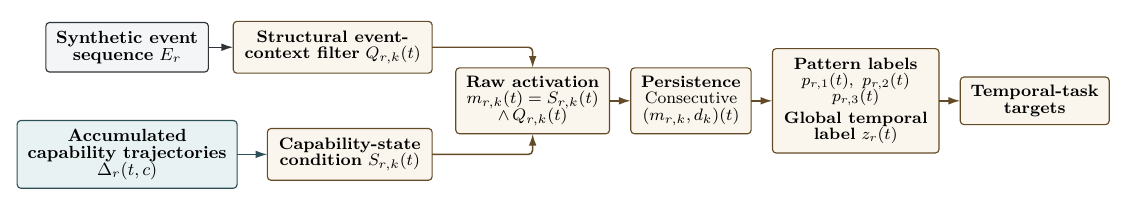}
    \caption{Offline construction of temporal-task targets. Capability-state conditions $S_{r,k}(t)$ and structural context filters $Q_{r,k}(t)$ define raw activations, which are converted into persistent pattern labels $p_{r,k}(t)$ and the global label $z_r(t)$. The annotations use complete synthetic trajectories, including future recovery and retrospective persistence. Primary predictors use the stated task windows; the separate shuffled pattern-detection diagnostic is qualified in Section~\ref{subsec:ablations-diagnostics}.}
    \label{fig:offline-target-construction}
\end{figure*}

Pattern labels follow the structural procedure formalized in Section~\ref{subsec:structural-convergence}. For each $k\in\mathcal{K}$, the procedure combines a capability-state condition over $\Delta_r(t,\cdot)$, a structural filter over $E_r$, and a persistence duration $d_k$. Figure~\ref{fig:offline-target-construction} summarizes the construction.

The structural filter evaluates adverse pressure, cross-domain diversity, distinct targets, target-graph connectedness, location-cluster coverage, and future supportive recovery evidence. The resulting labels are $p_{r,k}(t)$, $z_r(t)$, and $y_r^{\mathrm{sem}}$. Materialized target names are \texttt{target\_\allowbreak z}, \texttt{target\_\allowbreak p1}, \texttt{target\_\allowbreak p2}, and \texttt{target\_\allowbreak p3}; \texttt{run\_\allowbreak classification} uses \texttt{run\_\allowbreak label}.

Recovery checks and retrospective persistence make the temporal labels offline structural annotations. Subsequent observations can contribute to the annotation indexed at $t$, even though future state rows are not supplied to the primary predictor for that index. The whole-run shuffled pattern-detection condition is a distinct corruption diagnostic and does not preserve nominal window content.

\subsection{Positive and Negative Runs}
\label{subsec:positive-negative-runs}

Positive runs satisfy $y_r^{\mathrm{sem}}=1$, whereas negative runs satisfy $y_r^{\mathrm{sem}}=0$. 
Positive and negative branches are compiled jointly from a shared family template and ordered timestamp schedule. The positive branch must activate its prescribed patterns, whereas the negative branch must remain non-convergent and satisfy family-specific required and forbidden conditions. The negative branch is therefore not a single-event causal ablation of its positive counterpart.
Both classes may contain disruptive events; the benchmark contracts determine their distinction.

\subsection{Dataset Construction and Split Policy}
\label{subsec:dataset-construction}

TRACTA contains 200 runs, evenly divided between positive and negative labels. Each run has horizon $H=96$. Temporal datasets use $w=8$, with $\ell\in\{3,6,12\}$ for \texttt{early\_\allowbreak warning}, yielding 49,200 windows. The split contains 140 training, 30 validation, and 30 test runs and is performed by \texttt{run\_\allowbreak id}, preventing windows from the same run from crossing partitions.

The five scenario families each contain 20 positive and 20 negative runs. Within each family-label stratum, 14/3/3 runs enter training/validation/test. Thus, every family appears in every partition. Moreover, paired positive/negative runs can share an ordered timestamp schedule across partitions. The protocol evaluates held-out run instances from represented families; it does not establish independence between related pairs or generalization to unseen families. Table~\ref{tab:dataset-construction} summarizes the resulting dataset and split parameters.

\begin{table}[h]
\centering
\caption{Summary of TRACTA dataset construction.}
\label{tab:dataset-construction}
\footnotesize
\begin{tabular}{ll}
\hline
\textbf{Property} & \textbf{Value} \\
\hline
Runs (positive/negative) & 200 (100/100) \\
Run horizon & 96 \\
Temporal window size & 8 \\
Forecasting leads & 3, 6, 12 \\
Early-warning windows & 49,200 \\
Train/validation/test runs & 140/30/30 \\
Split policy & By \texttt{run\_\allowbreak id} \\
\hline
\end{tabular}
\end{table}

\subsection{Dataset Views and Shortcut Control}
\label{subsec:dataset-views-shortcut-control}

The full raw view retains global target and location identifiers for traceability, semantic resolution, and diagnostics, but is not a primary model input. The primary raw view replaces them with run-local aliases, removing stable cross-run identities while preserving recurrence within a run. The coarse view adds coarse entity descriptors. 
The sparse no-entity view retains only timestamp, event type, and subtype; because it also removes actor, confidence, and severity, it does not isolate entity removal alone.
The capability-trajectory view contains $D_r(t,c)$ and $\Delta_r(t,c)$ and is used by the semantic configurations. Table~\ref{tab:dataset-views} summarizes these dataset views and their roles.

\begin{table*}[h]
\centering
\caption{Dataset views and their roles.}
\label{tab:dataset-views}
\footnotesize
\begin{tabular}{p{0.24\textwidth}p{0.38\textwidth}p{0.30\textwidth}}
\hline
\textbf{View} & \textbf{Content} & \textbf{Role} \\
\hline
Full raw event & Global target and location identifiers & Traceability, semantic resolution, and diagnostics only. \\
Run-local aliased & Raw fields with local target/location aliases & Primary raw-event input. \\
Run-local coarse & Aliases plus coarse entity descriptors & Diagnostic/ablation view. \\
Sparse no-entity & Timestamp, event type, and subtype only & Sparse surface diagnostic. \\
Capability trajectories & Contextual direct impacts and accumulated states & Semantic rule and recurrent inputs. \\
\hline
\end{tabular}
\end{table*}

Run-local aliasing removes the direct cross-run identity channel, but within-run recurrence, paired-run structure, temporal organization, scalar fields, and other generator-dependent statistics may remain predictive. The diagnostics in Section~\ref{subsec:ablations-diagnostics} cover selected views and tasks and do not determine which features the main models use.

\section{Experimental Setup}
\label{sec:experimental-setup}

This section defines the protocol used to evaluate TRACTA. Section~\ref{sec:methodology} introduced benchmark construction, semantic trajectories, offline structural annotations, dataset views, and shortcut-control design. Here, we specify task execution, reference configurations, diagnostics, training, metrics, uncertainty summaries, and reproducibility assets.

\subsection{Benchmark Scope and Split Policy}
\label{subsec:experimental-scope}

All experiments use the benchmark described in Section~\ref{subsec:dataset-construction}: 200 synthetic runs with horizon $H=96$, temporal-window size $w=8$, and early-warning leads $\ell\in\{3,6,12\}$. The train--validation--test partition is defined by \texttt{run\_\allowbreak id}, preventing windows from the same run from crossing partitions. The frozen split contains 140 training, 30 validation, and 30 test runs, with positive/negative balance 70/70, 15/15, and 15/15, respectively.

The split is family-stratified by \texttt{family\_\allowbreak id} and \texttt{run\_\allowbreak label}; all five families occur in every partition. All 100 positive/negative pairs with the same family and numeric suffix share an ordered timestamp schedule, although other event fields differ. Forty-one pairs cross partitions, including 19 spanning training and test. The protocol therefore evaluates held-out runs from represented families, not independence between related pairs or generalization to unseen families.

\subsection{Tasks and Evaluation Hierarchy}
\label{subsec:experimental-tasks}

The experiments follow Section~\ref{subsec:task-hierarchy}. \texttt{early\_\allowbreak warning} and \texttt{pattern\_\allowbreak detection} are the primary temporal tasks.
They evaluate future and reference-time prediction of offline structural annotations.
\texttt{run\_\allowbreak classification} is a complementary full-run endpoint. Table~\ref{tab:task-hierarchy} summarizes the three tasks and their interpretive roles.

For \texttt{early\_\allowbreak warning}, the model observes a window ending at $t_{\mathrm{ref}}$ and predicts \texttt{target\_z}, \texttt{target\_p1}, \texttt{target\_p2}, and \texttt{target\_p3} at $t_{\mathrm{ref}}+\ell$. Valid reference times are $t_{\mathrm{ref}}\in\{w-1,\ldots,H-1-\ell\}$. The materialized corpus contains 49,200 windows: 17,200, 16,600, and 15,400 for leads 3, 6, and 12.

For \texttt{pattern\_\allowbreak detection}, the targets are indexed at $t_{\mathrm{ref}}$, with $t_{\mathrm{ref}}\in\{w-1,\ldots,H-1\}$, yielding 89 windows per run and 17,800 overall. For \texttt{run\_\allowbreak classification}, the model receives the complete run and predicts \texttt{run\_\allowbreak label}.

\begin{table*}[t]
\centering
\caption{TRACTA task hierarchy and interpretive role.}
\label{tab:task-hierarchy}
\footnotesize
\begin{tabular}{@{}p{0.20\textwidth}p{0.24\textwidth}p{0.30\textwidth}p{0.18\textwidth}@{}}
\toprule
\textbf{Task} & \textbf{Unit} & \textbf{Prediction target} & \textbf{Role} \\
\midrule
\texttt{early\_\allowbreak warning} & Window ending at $t_{\mathrm{ref}}$ & Offline label state at $t_{\mathrm{ref}}+\ell$, $\ell\in\{3,6,12\}$ & Primary; future annotation prediction. \\
\texttt{pattern\_\allowbreak detection} & Window ending at $t_{\mathrm{ref}}$ & Offline annotation indexed at $t_{\mathrm{ref}}$ & Primary; reference-time annotation prediction. \\
\texttt{run\_\allowbreak classification} & Complete run & Binary \texttt{run\_\allowbreak label} & Complementary endpoint. \\
\bottomrule
\end{tabular}
\end{table*}

\subsection{Input Views and Reference Configurations}
\label{subsec:model-configurations}

Raw-event neural references use the run-local aliased event view. Semantic-input configurations use contextual direct-impact and accumulated capability trajectories produced by the semantic reference layer. For temporal tasks, raw inputs contain events occurring within the eight-step window, whereas the first accumulated state in a semantic window may retain effects from earlier run history. Semantic inputs also use global-entity resolution and the sampled impact realization involved in target construction. Equal window indices therefore do not imply equal historical information or access to knowledge.

The comparison contains five configurations. \texttt{simple\_\allowbreak neural} is an MLP over padded raw-event sequences. \texttt{advanced\_\allowbreak neural} uses a bidirectional GRU~\cite{cho_gru_2014}. \texttt{transformer\_\allowbreak raw\_\allowbreak event} is an additional attention-based raw-event reference~\cite{vaswani_attention_2017,zerveas_time_series_transformer_2021}. \texttt{semantic\_\allowbreak baseline} is a contract-lite rule comparator. 
\texttt{neuro\_\allowbreak symbolic} is the repository name of the semantic-input recurrent reference, a bidirectional GRU over concatenated $D_r$ and $\Delta_r$ trajectories.
The comparison is configuration-level because input information, architecture, parameterization, training, and checkpoint selection are not fully matched. Table~\ref{tab:model-configurations} summarizes the five reference configurations and their roles.

\begin{table*}[t]
\centering
\caption{Reference configurations in the frozen TRACTA comparison.}
\label{tab:model-configurations}
\footnotesize
\begin{tabular}{@{}p{0.23\textwidth}p{0.27\textwidth}p{0.28\textwidth}p{0.14\textwidth}@{}}
\toprule
\textbf{Configuration} & \textbf{Input view} & \textbf{Architecture} & \textbf{Role} \\
\midrule
\texttt{simple\_\allowbreak neural} & Run-local aliased events & MLP over padded sequences & Minimal raw reference \\
\texttt{advanced\_\allowbreak neural} & Run-local aliased events & Bidirectional GRU & Recurrent raw reference \\
\texttt{transformer\_\allowbreak raw\_\allowbreak event} & Run-local aliased events & Transformer encoder & Attention-based raw reference \\
\texttt{semantic\_\allowbreak baseline} & Capability-trajectory dataset & Contract-lite rules & Rule reference \\
\texttt{neuro\_\allowbreak symbolic} & Capability-trajectory dataset & BiGRU over $D_r$ and $\Delta_r$ & Semantic-input recurrent reference \\
\bottomrule
\end{tabular}
\end{table*}

\subsection{Semantic-Input and Shortcut Diagnostics}
\label{subsec:ablations-diagnostics}

The component-zeroing diagnostics set either the 20 direct-impact channels or the 20 accumulated-capability channels to zero while retaining the 40-dimensional input and recurrent training defaults. They test conditional signal utility within the semantic-input configuration, not an isolated representation effect.

The archived \texttt{shuffled\_\allowbreak semantic} transformation differs by task: early-warning rows are permuted within each window; run-classification rows within the complete run; and pattern-detection rows across all 96 time steps before windows are extracted at their original indices. The pattern-detection condition changes both order and window content and may introduce originally future states. It is therefore a whole-run corruption diagnostic, not an order-only ablation.

Shortcut diagnostics compare the full global-identifier surface, run-local aliases, coarse and sparse views, scalar summaries, and alias recurrence. We executed standardized nearest-centroid surface diagnostics for run classification and early warning; recurrence diagnostics cover run classification only. No corresponding pattern-detection surface diagnostic or temporal-task recurrence control was executed. These diagnostic models do not reveal which features the main neural references use.

\subsection{Training and Checkpoint Selection}
\label{subsec:training-checkpoint-selection}

All trainable configurations use Adam, binary cross-entropy with logits, batch size 32, learning rate $10^{-3}$, and weight decay 0~\cite{kingma_adam_2015}. We apply positive-class weighting when both classes occur. Experiments use seeds 2026, 2027, and 2028. Raw MLP and raw BiGRU runs use at most 18 epochs with patience 5; the Transformer, semantic BiGRU, and semantic-input diagnostics use at most 24 epochs with patience 8.

Checkpoint selection uses validation macro-F1 by default. For semantic-input \texttt{run\_\allowbreak classification}, it uses mean family macro-F1, with global macro-F1 as tie-breaker. We select the contract-lite rule profile on the training data before validation and test evaluation.

The recurrent references are not capacity- or training-matched. The raw BiGRU has 110 input features, a hidden size of 96 per direction, and 120,001 parameters; the semantic BiGRU has 40 channels, a hidden size of 128, and 130,817 parameters. Their epoch budgets, patience, and run-classification checkpoint criteria also differ. Table~\ref{tab:recurrent-comparison} summarizes the corresponding recurrent-reference configurations.

\begin{table*}[t]
\centering
\caption{Frozen recurrent reference configurations.}
\label{tab:recurrent-comparison}
\footnotesize
\begin{tabular}{lcc}
\toprule
\textbf{Property} & \textbf{Raw BiGRU} & \textbf{Semantic BiGRU} \\
\midrule
Input dimension & 110 & 40 \\
Hidden size per direction & 96 & 128 \\
Trainable parameters & 120,001 & 130,817 \\
Maximum epochs / patience & 18 / 5 & 24 / 8 \\
EW actual epochs, range / median & 6--18 / 15 & 9--24 / 19 \\
PD actual epochs, range / median & 8--18 / 10 & 10--24 / 16.5 \\
RC actual epochs, range / median & 11--15 / 15 & 17--24 / 22 \\
RC validation selection & Global macro-F1 & Mean family macro-F1 \\
\bottomrule
\end{tabular}
\end{table*}

We fit raw categorical vocabularies on the training data. Each raw event has 110 encoded features; we exclude run, family, event identifiers, and notes. We pad raw temporal and full-run sequences to 24 and 96 event positions. Semantic inputs contain 40 unnormalized channels and eight state rows per temporal window. Alias dictionaries use the complete run entity set before window extraction.

\subsection{Evaluation Metrics and Uncertainty Summaries}
\label{subsec:evaluation-metrics}

Macro-F1 is the primary metric~\cite{sokolova_measures_2009,he_imbalanced_2009}. Let $\mathcal{Y}=\{z,p_1,p_2,p_3\}$ and $\mathcal{L}=\{3,6,12\}$. The headline temporal scores are
\begin{equation}
\mathrm{MacroF1}_{\mathrm{EW}}=
\frac{1}{|\mathcal{Y}||\mathcal{L}|}
\sum_{y\in\mathcal{Y}}\sum_{\ell\in\mathcal{L}}\mathrm{MacroF1}(y,\ell),
\end{equation}
\begin{equation}
\mathrm{MacroF1}_{\mathrm{PD}}=
\frac{1}{|\mathcal{Y}|}\sum_{y\in\mathcal{Y}}\mathrm{MacroF1}(y).
\end{equation}

The pipeline computes grouped-bootstrap bounds from fixed saved predictions using 1,000 resamples by \texttt{run\_\allowbreak id}. Resampling is performed separately for each seed and setting, after which we average interval endpoints. The reported bounds are descriptive uncertainty summaries, not joint confidence intervals for the headline aggregates or tests of statistically resolved superiority. They exclude alternative splits, retraining, and benchmark regeneration.

\subsection{Reproducibility Assets}
\label{subsec:reproducibility-assets}

The workflow records resolved configurations, split assignments, task definitions, saved predictions, aggregate exports, diagnostics, bootstrap artifacts, and manifest metadata. Appendix~\ref{app:reproducibility} identifies the frozen release and execution; Supplementary Material S1 provides artifact-level provenance and regeneration checks.

\section{Results}
\label{sec:results}

This section reports the TRACTA results according to the task hierarchy and evaluation protocol defined in Section~\ref{sec:experimental-setup}. The analysis addresses three questions: how the evaluated semantic-input recurrent reference compares with the raw-event and rule-based configurations; what the diagnostic variants reveal about the semantic inputs; and which shortcut-related signals remain after run-local aliasing. Macro-F1 is the primary metric.

\subsection{Aggregate Comparison}
\label{subsec:aggregate-results}

Table~\ref{tab:main-benchmark-results} reports the aggregate macro-F1 point estimates. The \texttt{neuro\_\allowbreak symbolic} configuration has the highest headline estimate on all three tasks.

The largest aggregate margins occur on the temporal tasks. For \texttt{early\_\allowbreak warning}, the semantic-input recurrent reference reaches 0.6860, compared with 0.6152 for the strongest raw-event reference. For \texttt{pattern\_\allowbreak detection}, the corresponding values are 0.7849 and 0.6541. The \texttt{run\_\allowbreak classification} margin is smaller: 0.7865 versus 0.7550.

Adding the tested Transformer reference does not change the aggregate ranking. This broadens the raw-event comparison, but does not isolate representation as the cause of the observed differences, because the evaluated configurations differ in architecture, parameterization, training budget, checkpoint selection, and available information.
Target- and lead-level decompositions are reported in Appendix~\ref{app:extended-results}.

\begin{table*}[t]
\centering
\caption{Aggregate macro-F1 point estimates on TRACTA. Early warning is averaged across four targets and three leads; pattern detection is averaged across four targets; run classification uses the binary run label.}
\label{tab:main-benchmark-results}
\footnotesize
\begin{tabular}{lccc}
\hline
\textbf{Configuration} & \textbf{Early warning} & \textbf{Pattern detection} & \textbf{Run classification} \\
\hline
\texttt{neuro\_\allowbreak symbolic} & \textbf{0.6860} & \textbf{0.7849} & \textbf{0.7865} \\
\texttt{advanced\_\allowbreak neural} & 0.6152 & 0.6541 & 0.7424 \\
\texttt{transformer\_\allowbreak raw\_\allowbreak event} & 0.6145 & 0.6432 & 0.6931 \\
\texttt{simple\_\allowbreak neural} & 0.6146 & 0.6302 & 0.7550 \\
\texttt{semantic\_\allowbreak baseline} & 0.5948 & 0.6378 & 0.5928 \\
\hline
\end{tabular}
\end{table*}

\subsection{Temporal Tasks}
\label{subsec:results-temporal}

For \texttt{early\_\allowbreak warning}, the semantic-input recurrent reference achieves the highest score after aggregation across targets and leads. Raw-event references remain predictive, and the granular decomposition contains four target--lead settings led by comparator configurations. The result therefore describes the aggregate ranking, not uniform dominance or prospective alert performance.

For \texttt{pattern\_\allowbreak detection}, the semantic-input recurrent reference has the largest aggregate margin. This configuration receives accumulated capability and direct-impact trajectories constructed in the same modeled state space used by the offline annotation procedure. The result is therefore consistent with the benchmark's knowledge alignment, but does not show that the fitted model recovers the intended structural mechanism or that semantic representation alone causes the difference.

Both temporal tasks predict offline annotations generated from complete trajectories. In particular, \texttt{early\_\allowbreak warning} predicts a future label state rather than exclusively a new onset, while \texttt{pattern\_\allowbreak detection} predicts the annotation indexed at the reference time rather than an independently established online state.

\subsection{Run Classification}
\label{subsec:results-run-classification}

For \texttt{run\_\allowbreak classification}, the semantic-input recurrent reference has the highest point estimate, but the two non-Transformer neural references are close. Complete runs expose aggregate and recurrence-related signals not specific to the temporal annotation mechanism. The task is therefore retained as a complementary endpoint rather than as the main evidence for temporal structural reasoning.

\subsection{Semantic-Input Diagnostic Variants}
\label{subsec:results-ablation}

Table~\ref{tab:ablation-results} reports the archived semantic-input diagnostic variants. Zeroing either accumulated capability trajectories or contextual direct-impact trajectories lowers the point estimate on each task relative to the full configuration.

The archived shuffled condition is task-dependent. Early-warning windows are permuted internally, full runs are permuted for run classification, and pattern detection shuffles the complete 96-step sequence before windows are extracted at their original indices. The pattern-detection variant therefore changes both order and window content and may place originally future states into a window. Its large decrease reflects this whole-run corruption condition, not an isolated causal estimate of temporal ordering. The diagnostic suite shows that both semantic channel groups contain useful signal, but does not uniquely decompose the full configuration's advantage.

\begin{table*}[t]
\centering
\caption{Aggregate macro-F1 point estimates for the semantic-input diagnostic variants. The shuffled condition is task-dependent and is not a uniform order-only ablation.}
\label{tab:ablation-results}
\footnotesize
\begin{tabular}{lccc}
\hline
\textbf{Configuration} & \textbf{Early warning} & \textbf{Pattern detection} & \textbf{Run classification} \\
\hline
\texttt{neuro\_\allowbreak symbolic} & \textbf{0.6860} & \textbf{0.7849} & \textbf{0.7865} \\
\texttt{capability\_\allowbreak only} & 0.6633 & 0.7412 & 0.7166 \\
\texttt{direct\_\allowbreak impact\_\allowbreak only} & 0.6614 & 0.7358 & 0.6686 \\
\texttt{shuffled\_\allowbreak semantic} & 0.6754 & 0.5422 & 0.7310 \\
\hline
\end{tabular}
\end{table*}

Complete differences and diagnostic implementation details are reported in Appendix~\ref{app:ablation-diagnostics}.

\subsection{Shortcut Diagnostics}
\label{subsec:results-leakage}

The full raw global-identifier diagnostic reaches 0.8990 macro-F1 on \texttt{run\_\allowbreak classification}, confirming that stable target and location identities provide a strong predictive channel when exposed directly. The run-local aliased surface reaches 0.5623, while target, location, and combined alias-recurrence summaries reach 0.5662, 0.6997, and 0.7321, respectively.

Run-local aliasing removes stable global identities by construction, but the diagnostics show that shallow predictivity remains, particularly through recurrence summaries. These values do not demonstrate that the primary neural models use the same features. Moreover, we ran recurrence diagnostics only for run classification, and no corresponding pattern-detection diagnostic is available.

The complete diagnostic table is reported in Appendix~\ref{app:ablation-diagnostics}.

\subsection{Uncertainty and Setting-Level Heterogeneity}
\label{subsec:results-uncertainty}

Grouped-bootstrap uncertainty bounds are reported as descriptive context. The pipeline bootstraps fixed predictions separately by seed and task setting and then summarizes the resulting bounds; it does not jointly bootstrap each headline task aggregate or capture alternative splits, retraining, or benchmark regeneration. The broad overlap therefore does not support statistically resolved superiority between configurations.

The semantic-input recurrent reference has the highest headline point estimate on all three tasks, but four early-warning target--lead settings are led by comparator configurations. These exceptions and the uncertainty summaries are reported in Appendix~\ref{app:bootstrap-analyses}.

\subsection{Summary}
\label{subsec:results-summary}

In summary, the semantic-input recurrent reference has the highest aggregate point estimates in the frozen comparison. The evidence is configuration-level: it is qualified by knowledge alignment, unmatched configuration factors, offline target construction, residual shallow predictivity, paired-run dependence, setting-level exceptions, and descriptive rather than inferential uncertainty bounds.

\section{Discussion}
\label{sec:discussion}

The semantic-input recurrent reference has the highest aggregate point estimates in the frozen TRACTA comparison, with the largest margins on the two temporal tasks. This is a configuration-level finding under the benchmark's knowledge alignment, input policies, architectures, training settings, and offline target construction; it is not evidence of a general or causal advantage of semantic representation.

\subsection {Configuration-Level Interpretation}

The semantic-input recurrent reference operates on contextual direct-impact and accumulated capability trajectories produced by the semantic reference layer. These variables are aligned with the capability space used to construct the structural annotations, whereas raw-event references must learn from the aliased event surface. This alignment may contribute to the observed gap. However, the recurrent configurations also differ in available historical information, input dimensionality, hidden size, parameter count, training budget, and checkpoint selection. The experiments therefore do not isolate representation as the causal factor.

\subsection{Temporal Tasks and Structural Convergence}

The temporal tasks operationalize temporal structural reasoning as prediction of offline annotations. For \texttt{early\_\allowbreak warning}, the semantic-input recurrent reference has the highest aggregate point estimate, but comparator configurations lead in four target--lead settings. The task predicts a future label state, not exclusively the onset of a new episode. For \texttt{pattern\_\allowbreak detection}, the larger aggregate margin is compatible with the alignment between semantic inputs and the annotation space, but does not establish reconstruction of the full structural contract or online detection capability.

By contrast, \texttt{run\_\allowbreak classification} operates on complete runs and may rely more heavily on aggregate or recurrence-related signals. This may help explain the smaller gap and supports its interpretation as a complementary endpoint. The grouped-bootstrap uncertainty bounds overlap, particularly for this task, and do not support statistically resolved superiority. Detailed setting-level results are reported in Appendices~\ref{app:extended-results} and~\ref{app:bootstrap-analyses}.

\subsection{Raw-Event and Rule-Based References}

Raw-event references remain predictive, especially on \texttt{run\_\allowbreak classification}. The tested Transformer broadens the reference suite but cannot rule out stronger architectures or alternative optimization choices. The strongest raw-event configuration also varies by task, showing that greater architectural complexity does not guarantee a higher score under the frozen protocol. These observations qualify the aggregate ranking and do not establish that event-level information is intrinsically less suitable.

The \texttt{semantic\_\allowbreak baseline} is a restricted rule reference selected from fixed profiles using training data. Its lower aggregate point estimates characterize that comparator and its task inputs. Because the offline annotations are themselves rule-defined, this comparison cannot establish an intrinsic limitation of symbolic reasoning or identify which temporal evidence the recurrent configuration exploits. Stronger symbolic, probabilistic, temporal-logic, or graph-based systems remain unevaluated.

\subsection{Semantic-Input Diagnostics}

The component-zeroing variants have lower aggregate point estimates when either capability trajectories or contextual direct-impact trajectories are unavailable. They show that each channel group contains useful information within the evaluated semantic-input configuration, but do not estimate an interaction effect or robustness to misspecified semantic knowledge.

The shuffled condition cannot be interpreted uniformly across tasks. In particular, the pattern-detection implementation permutes the entire trajectory before extracting windows at their original indices, changing both temporal order and window content and potentially introducing originally future states. Its large decrease therefore characterizes this whole-run corruption condition rather than an isolated temporal-order effect.

\subsection{Shortcut Diagnostics and Dependence Structure}

Run-local aliasing removes stable global target and location identities from the primary raw input by construction. The executed surface diagnostics nevertheless retain predictive signal, and the run-classification recurrence summaries are especially informative. These diagnostics do not show which cues the main models use; recurrence diagnostics are unavailable for the temporal tasks, and no corresponding pattern-detection surface diagnostic was executed. In addition, positive/negative pairs may share ordered timestamp schedules across partitions. The extent to which residual surface statistics and related-run dependence contribute to the reported scores therefore remains unresolved.

\subsection{Implications for Benchmark and Neuro-Symbolic Design}

TRACTA illustrates how a semantic reference layer can provide an inspectable interface between heterogeneous events, modeled capability states, rule-based annotations, and temporal predictors. Its principal contribution is the benchmark structure and accompanying diagnostics, not evidence that the evaluated recurrent architecture is generally superior. The next methodological step is to test this interface under perturbed semantic mappings, paired-group or family-held-out splits, broader reference models, and data conditions not generated by the same knowledge-aligned process.

\subsection{Summary}

Overall, the frozen comparison places the semantic-input recurrent reference first in aggregate point estimates. The result remains conditional on knowledge alignment, unmatched configuration factors, offline labels, incomplete shortcut diagnostics, related-run dependence, and descriptive uncertainty summaries.

\section{Limitations and Threats to Validity}
\label{sec:limitations}

This section identifies the conditions that bound the interpretation of TRACTA. The reported results concern a fixed, knowledge-aligned synthetic benchmark and should not be read as operational validation or as a causal comparison of representation families.

\subsection{Synthetic Setting and External Validity}
\label{subsec:limitations-synthetic}

The primary limitation is TRACTA's synthetic nature. Synthetic construction enables controlled evaluation under explicit assumptions, but does not establish real-world validity~\cite{nikolenko_synthetic_data_2021,paullada_dataset_development_2021}. Operational data may contain incomplete observations, attribution uncertainty, distribution shifts, heterogeneous reporting, and human decision processes not represented here.

The benchmark uses MDO-like event schemas, capability dependencies, entity structures, and pattern contracts. The authors have not evaluated transfer to other domains, alternative ontologies, or independently generated data. The evidence is therefore methodological and benchmark-specific.

\subsection{Knowledge Alignment and Configuration Comparability}
\label{subsec:limitations-semantic}

A central construct-validity limitation is the alignment between the semantic inputs and the target-generation process. Accumulated capability trajectories contribute to both the semantic-input configurations and the offline structural annotations. The resulting comparison tests access to benchmark-modeled concepts, not whether semantic representations are generally superior. Robustness to noisy, incomplete, perturbed, or misspecified semantic mappings remains untested.

The evaluated configurations are also not fully matched. The raw-event and semantic-input recurrent references differ in input dimensionality, available historical information, hidden size, parameter count, training duration, and checkpoint selection. The observed gaps are therefore configuration-level findings and cannot be attributed uniquely to representation, architecture, optimization, or knowledge access.

\subsection{Offline Targets and Task Interpretation}
\label{subsec:limitations-task-interpretation}

The temporal targets are offline annotations generated from complete trajectories. Future recovery checks and retrospective persistence can affect the label indexed at time $t$. Consequently, \texttt{pattern\_\allowbreak detection} is not a validation of independently established online state detection, and \texttt{early\_\allowbreak warning} predicts a future label state rather than exclusively the onset of a new episode. Aggregate results also mask target--lead heterogeneity, including settings led by comparator configurations.

\subsection{Corpus Dependence, Split Policy, and Uncertainty}
\label{subsec:limitations-size}

The benchmark contains 200 runs, but they should not be treated as 200 independent scenario realizations. Runs are generated as positive/negative pairs from shared family templates and ordered timestamp schedules, and paired branches may cross train, validation, and test partitions. The run-level split prevents windows from the same run from crossing partitions, but does not enforce paired-group separation or family-held-out evaluation. All five families occur in every partition.

The independent support is therefore smaller and more structured than the temporal-window count suggests. Results may change under paired-group splits, family-held-out evaluation, alternative partitions, or new benchmark generations. We compute the grouped-bootstrap bounds from fixed saved predictions separately by seed and setting, then summarize them. They do not jointly bootstrap the headline task aggregates or capture split, retraining, and generation uncertainty, and they do not demonstrate statistically resolved differences between configurations.

\subsection{Diagnostic Scope}
\label{subsec:limitations-shortcuts}

Run-local aliasing removes stable global target and location identities from the primary raw input by construction, but does not eliminate shallow predictivity. Surface diagnostics are available only for selected tasks, recurrence diagnostics are limited to \texttt{run\_\allowbreak classification}, and no corresponding pattern-detection diagnostic was executed. Diagnostic performance also does not reveal which features the main neural models use. Residual recurrence, scalar fields, temporal schedules, paired-run structure, and other generator-dependent cues therefore remain unresolved validity concerns.

The semantic-input diagnostics have related limitations. Component-zeroing establishes that each trajectory block contains useful signal within the tested configuration, but does not estimate an interaction effect. The shuffled condition is task-dependent; for pattern detection it changes both order and window content and can include originally future states. It should therefore be interpreted as a corruption diagnostic, not as an isolated test of temporal order.

\subsection{Model and Reproducibility Scope}
\label{subsec:limitations-model}

The conclusions are limited to the tested model space: three raw-event neural references, a contract-lite semantic rule comparator, and one recurrent semantic-input configuration. Larger attention-based models, graph architectures, stronger symbolic systems, alternative temporal models, and better-matched representation comparisons may produce different results.

The frozen configurations, predictions, splits, and provenance artifacts improve traceability, but exact replication remains dependent on software, hardware, preprocessing, and repository state. Distribution of synthetic operational artifacts may also remain subject to applicable institutional and sensitivity constraints.

\subsection{Summary}

Overall, TRACTA supports a bounded comparison of the archived configurations. Stronger conclusions require independently generated or real data, perturbed semantic mappings, paired-group and family-held-out splits, matched model comparisons, broader diagnostics, and inferential uncertainty analyses defined at the headline-task level.

\section{Conclusions and Future Work}
\label{sec:conclusions}

This paper introduced TRACTA, a knowledge-aligned synthetic benchmark for comparing raw-event, contract-lite rule, and semantic-input recurrent references on offline temporal structural annotations. In the frozen comparison, the semantic-input recurrent reference has the highest aggregate point estimates, with the largest margins on the two temporal tasks. This configuration-level result does not isolate a causal representation effect and is qualified by overlapping uncertainty bounds, target--lead heterogeneity, and differences in available information, architecture, and training.

The main contribution is methodological: TRACTA makes explicit the event-to-capability mapping, contract-based offline target construction, input views, reference configurations, diagnostics, and frozen evidence used in the comparison.

Future work should prioritize paired-group and family-held-out evaluation, independently generated or real data, perturbation of the semantic reference layer, better-matched configurations, broader shortcut diagnostics, and uncertainty analyses defined directly over the headline task aggregates.

TRACTA therefore provides a reproducible basis for studying knowledge-aligned temporal prediction, while leaving operational validity and general representation-level conclusions open.

\section*{CRediT authorship contribution statement}

\noindent\textbf{Michael Romei de Socio:}
Conceptualization, Methodology, Software, Writing -- original draft.

\noindent\textbf{Gian Luca Pozzato:}
Writing -- review and editing, Supervision, Resources.

\noindent\textbf{Alessio Merlo:}
Writing -- review and editing, Supervision, Project administration, Methodology.

\section*{Acknowledgements}
The authors gratefully acknowledge the ``HPC4AI'' initiative~\cite{aldinucci_hpc4ai_2018} for providing access to high-performance computing resources that supported the preliminary implementation of this work.

\section*{Funding}
This research received no specific grant from funding agencies in the public, commercial, or not-for-profit sectors.

\section*{Declaration of Competing Interest}
The authors declare no known competing financial interests or personal relationships that could have influenced the work reported in this paper.

\section*{Data Availability}
The TRACTA v1.0.0 benchmark data, split definitions, and archived experimental outputs supporting this study are available on Zenodo at \url{https://doi.org/10.5281/zenodo.22715064}.

\section*{Code Availability}
The source code and frozen configurations used for benchmark construction, semantic and task materialization, model training, evaluation, and regeneration of the reported analyses are available in the same Zenodo record.

\section*{Supplementary Material}
Supplementary Material S1 contains additional documentation
on reproducibility, execution environment, and artifact traceability.

\clearpage
\section*{Appendices}

\appendix

\numberwithin{equation}{section}
\numberwithin{table}{section}

\section{Pattern Formalization and Label Construction}
\label{app:pattern-formalization}

This appendix specifies the operational construction of the semantically grounded trajectories and offline structural annotations used in TRACTA. It complements Sections~\ref{sec:problem-framing} and~\ref{sec:methodology}; definitions introduced in the main text are referenced rather than duplicated.

\subsection{Run Representation and Capability Space}
\label{app:run-capability-space}

Using $E_r$ from Equation~\eqref{eq:run-sequence}, each event is represented as
\begin{equation}
\begin{aligned}
e=(&\texttt{event\_id},\tau,\texttt{event\_type},\texttt{subtype},\\
   &\texttt{target},\texttt{location},\texttt{actor},\\
   &\texttt{confidence},\texttt{severity},\texttt{notes}).
\end{aligned}
\end{equation}
The horizon is $H=96$, with $\mathcal{T}=\{0,\ldots,95\}$. The frozen semantic reference layer contains $|\mathcal{C}|=20$ capabilities, stored in the following order:
{\footnotesize
\begin{equation}
\begin{aligned}
\mathcal{C}=\{&\texttt{EnergyAvailability},\texttt{PortOperations},\\
&\texttt{LogisticsFlowNorth},\texttt{LogisticsC2Integrity},\\
&\texttt{MaritimeAccessIntegrity},\texttt{CoalitionSupportStability},\\
&\texttt{RoadCorridorResilience},\texttt{RailNetworkContinuity},\\
&\texttt{PortYardFluidity},\texttt{CargoTrackingIntegrity},\\
&\texttt{AirspaceAccessReliability},\texttt{FuelDistributionStability},\\
&\texttt{AidDeliveryRate},\texttt{LegislativeApprovalStability},\\
&\texttt{PublicNarrativeStability},\texttt{ExternalDiplomaticAlignment},\\
&\texttt{EnergyRepairReadiness},\texttt{PortRecoveryCapacity},\\
&\texttt{RailRepairCapacity},\texttt{CommandAdaptationCapacity}\}.
\end{aligned}
\end{equation}}
Contextual impact resolution uses \texttt{subtype}, \texttt{target}, \texttt{location}, and \texttt{severity}. Structural filtering additionally uses $\tau$, \texttt{event\_type}, induced location class, and the relevant entity relations. The fields \texttt{confidence}, \texttt{actor}, and \texttt{notes} are retained but are not used by the inspected label-generation path.

\subsection{Contextual Impact and Trajectory Materialization}
\label{app:direct-impact-resolution}

The contextual impact range is defined in Equation~\eqref{eq:contextual-impact-range}. The criticality multiplier is
\begin{equation}
\begin{aligned}
m_{\mathrm{crit}}(e)=\operatorname{clip}_{[0.7,1.3]}\Big(&1+0.6\big(
\mathrm{criticality}(\mathrm{target}(e))\\
&-0.5\big)\Big).
\end{aligned}
\end{equation}
The dislocation multiplier is $0.85$ when the event location falls outside the target's allowed locations and $1.0$ otherwise. Location effects combine class- and instance-level capability modifiers. The noise generator is initialized with seed 42 for each run's semantic materialization; this is distinct from the recorded run-construction seed schedule and the three model-training seeds. These constants define the frozen benchmark specification. No comprehensive sensitivity analysis is reported.

Let $(\tilde{\ell}_{e,c},\tilde{u}_{e,c})$ be the resolved range. For mitigating impacts,
\begin{equation}
b_{e,c}=\tilde{\ell}_{e,c}+\mathrm{severity}(e)
\big(\tilde{u}_{e,c}-\tilde{\ell}_{e,c}\big),
\end{equation}
and otherwise
\begin{equation}
b_{e,c}=\tilde{u}_{e,c}+\mathrm{severity}(e)
\big(\tilde{\ell}_{e,c}-\tilde{u}_{e,c}\big).
\end{equation}
The sampled impact is
\begin{equation}
x_{e,c}=\operatorname{clip}_{[-1,1]}(b_{e,c}+\epsilon),
\qquad \epsilon\sim\mathcal{U}[-0.01,0.01].
\end{equation}
Aggregating these values yields $D_r(t,c)$ in Equation~\eqref{eq:direct-impact-trajectory}. Individual impacts are clipped, but their per-timestamp sum is not. The accumulated trajectory follows Equations~\eqref{eq:capability-trajectory-initialization} and~\eqref{eq:capability-trajectory-recursion}. The frozen benchmark uses $\lambda=0.20$ and explicit weights for all 26 graph edges; the nominal $0.30$ fallback is inactive in the reported graph.

\subsection{Pattern Activation and Structural Filtering}
\label{app:structural-filter}

Pattern activation follows Equations~\eqref{eq:capability-state-condition}--\eqref{eq:semantic-run-label}. Repository fields \texttt{p1\_t}, \texttt{p2\_t}, \texttt{p3\_t}, and \texttt{z\_t} correspond to $p_{r,1}(t)$, $p_{r,2}(t)$, $p_{r,3}(t)$, and $z_r(t)$. The semantic run label $y_r^{\mathrm{sem}}$ and \texttt{run\_label} coincide for the frozen corpus, although \texttt{run\_classification} uses the latter.

For event $e$ and capability $c$, the deterministic adverse and supportive proxies are
\begin{equation}
\begin{aligned}
\mathrm{negative\_caps}_{e,c}={}&
\max\!\left(0,-\min(\tilde{\ell}_{e,c},\tilde{u}_{e,c})\right)\\
&\cdot\mathrm{severity}(e),
\end{aligned}
\end{equation}
\begin{equation}
\begin{aligned}
\mathrm{supportive\_caps}_{e,c}={}&
\max\!\left(0,\max(\tilde{\ell}_{e,c},\tilde{u}_{e,c})\right)\\
&\cdot\mathrm{severity}(e).
\end{aligned}
\end{equation}
They do not use sampled \texttt{direct\_values}. For pattern $k$, the look-back window is
\begin{equation}
W^-_{k,t}=\{\tau\in\mathcal{T}:t-\omega_k+1\leq\tau\leq t\},
\end{equation}
and accumulated adverse pressure is
\begin{equation}
A_{r,k,t}(c)=\sum_{\substack{e\in E_r:\\ \tau(e)\in W^-_{k,t}}}
\mathrm{negative\_caps}_{e,c}.
\end{equation}
The filter checks pattern-specific event counts, domain and target diversity, direct-pressure groups, target-graph connectedness, location-cluster coverage, and future supportive recovery evidence. If $U_{r,k,t}$ is the relevant target set,
\begin{equation}
\frac{|\operatorname{LCC}(U_{r,k,t})|}{|U_{r,k,t}|}\geq\rho_k,
\qquad U_{r,k,t}\neq\emptyset.
\end{equation}
The recovery window is
\begin{equation}
W^+_{k,t}=\{\tau\in\mathcal{T}:t<\tau\leq
\min(t+\gamma_k,t_{\max})\}.
\end{equation}
The primary predictors receive the stated observation windows, but the targets are offline annotations: future recovery affects $Q_{r,k}(t)$ and persistence retrospectively labels every position in a qualifying contiguous segment. This distinction precludes interpreting \texttt{pattern\_detection} as independently validated online detection.

\subsection{Pattern Contracts}
\label{app:pattern-contracts}

The capability-state conditions are reported in Table~\ref{tab:pattern-state-conditions}. Here, $\operatorname{AtLeast}_q$ means that at least $q$ listed predicates hold. Table~\ref{tab:pattern-structural-parameters} reports the corresponding base structural-contract parameters.

\begin{table*}[t]
\centering
\caption{Capability-state conditions. A pair $X{:}{-}a$ denotes $X\leq-a$.}
\label{tab:pattern-state-conditions}
\small
\renewcommand{\arraystretch}{1.22}
\begin{tabularx}{0.96\textwidth}{@{}c>{\raggedright\arraybackslash}X@{}}
\toprule
\textbf{Pattern} & \textbf{State condition} \\
\midrule
$P_1$ & $\operatorname{AtLeast}_2\{\mathrm{PO}{:}{-}.75,\mathrm{FDS}{:}{-}.80,\mathrm{RCR}{:}{-}.65,\mathrm{RNC}{:}{-}.60\}\land\operatorname{AtLeast}_1\{\mathrm{LFN}{:}{-}.75,\mathrm{ADR}{:}{-}.80\}\land\operatorname{AtLeast}_1\{\mathrm{PYF}{:}{-}.60,\mathrm{MAI}{:}{-}.45\}$ \\
$P_2$ & $(\mathrm{LC2I}{:}{-}.55\land\mathrm{CAC}{:}{-}.45)\land\operatorname{AtLeast}_1\{\mathrm{LFN}{:}{-}.75,\mathrm{PO}{:}{-}.70\}\land\operatorname{AtLeast}_1\{\mathrm{CTI}{:}{-}.10,\mathrm{RCR}{:}{-}.55,\mathrm{PYF}{:}{-}.60\}$ \\
$P_3$ & $(\mathrm{CSS}{:}{-}.55\land\mathrm{ADR}{:}{-}.75)\land\operatorname{AtLeast}_2\{\mathrm{LAS}{:}{-}.45,\mathrm{PNS}{:}{-}.08,\mathrm{EDA}{:}{-}.08\}$ \\
\bottomrule
\end{tabularx}
\end{table*}

\begin{table*}[t]
\centering
\caption{Base structural-contract parameters.}
\label{tab:pattern-structural-parameters}
\small
\setlength{\tabcolsep}{5pt}
\renewcommand{\arraystretch}{1.16}
\begin{tabularx}{0.96\textwidth}{@{}lcccccc>{\raggedright\arraybackslash}X@{}}
\toprule
\textbf{Pattern} & $d_k$ & $\omega_k$ & \textbf{Events} & \textbf{Domains} & \textbf{Targets} & $\rho_k$ & \textbf{Recovery gate} \\
\midrule
$P_1$ & 4 & 14 & 4 & 2 & 3 & 0.30 & $\gamma_1=8$; at most 2 supportive events and 0.22 supportive pressure \\
$P_2$ & 6 & 12 & 4 & 2 & 3 & 0.40 & $\gamma_2=6$; at most 2 supportive events and 0.16 supportive pressure \\
$P_3$ & 8 & 16 & 4 & 2 & 3 & 0.45 & $\gamma_3=8$; at most 2 supportive events and 0.16 supportive pressure \\
\bottomrule
\end{tabularx}
\end{table*}

All patterns require at least two location classes; base coverage thresholds are 0.65, 0.60, and 0.75 for $P_1$, $P_2$, and $P_3$. $P_1$ admits \texttt{port\_energy\_lock} and \texttt{corridor\_binding\_lock}; the latter overrides the connectedness ratio and location coverage to 0.40 and 0.70. Complete mode-specific groups and clusters are provided in the frozen contract file.

For each maximal contiguous interval of raw activation, every position receives label 1 if the interval length reaches $d_k$; otherwise every position receives 0. The operator therefore backfills a qualifying interval rather than delaying its first positive label by $d_k$ steps.

\FloatBarrier

\subsection{Task-Specific Targets and Worked Example}
\label{app:task-target-construction}

For \texttt{early\_warning},
\begin{equation}
\begin{aligned}
\texttt{target\_z}&=y_{r,z}(t_{\mathrm{ref}}+\ell),\\
\texttt{target\_pk}&=y_{r,k}(t_{\mathrm{ref}}+\ell),
\qquad k\in\mathcal{K},
\end{aligned}
\end{equation}
where $\ell\in\{3,6,12\}$. For \texttt{pattern\_detection},
\begin{equation}
\begin{aligned}
\texttt{target\_z}&=y_{r,z}(t_{\mathrm{ref}}),\\
\texttt{target\_pk}&=y_{r,k}(t_{\mathrm{ref}}),
\qquad k\in\mathcal{K}.
\end{aligned}
\end{equation}
Thus, \texttt{pattern\_detection} predicts the offline annotation indexed at the reference time.

As a frozen example, positive run 000 of family F1 contains three time-84 \nolinkurl{MissileStrikePortInfrastructure} events whose PortOperations impacts sum to approximately $D(84,\mathrm{PO})=-2.4578$. The required $P_1$ state and context conditions hold, and the persistent annotation is positive at times 84--91. Pattern detection at $t_{\mathrm{ref}}=84$ uses window 77--84; early warning at $t_{\mathrm{ref}}=81$ and lead 3 uses window 74--81 with the same target. This illustrates stored input/annotation pairs, not a claim about any model prediction.

\subsection{Implementation Boundary}
\label{app:implementation-notes}

The TRACTA v1.0.0 release uses the contract-based path described above. Changes to semantic files, graph weights, contracts, persistence, or recovery gates define a different benchmark specification and require separate documentation and evaluation.  

\section{Reproducibility and Artifact Traceability}
\label{app:reproducibility}

The experimental protocol and exported assets are described in Sections~\ref{sec:methodology} and~\ref{sec:experimental-setup}. This appendix identifies the frozen evidence underlying the reported results. Table~\ref{tab:reproducibility-identity} summarizes the frozen release and execution identity.

\begin{table}[htbp]
\centering
\caption{Reproducibility identity of the reported TRACTA experiments.}
\label{tab:reproducibility-identity}
\small
\renewcommand{\arraystretch}{1.12}

\begin{tabularx}{\linewidth}{
  @{}
  >{\raggedright\arraybackslash}p{0.34\linewidth}
  >{\raggedright\arraybackslash}X
  @{}
}
\toprule
\textbf{Item} & \textbf{Recorded identity} \\
\midrule
Benchmark release & TRACTA v1.0.0 \\
Reference execution & Frozen reference execution associated with TRACTA v1.0.0 \\
Software environment & Python 3.12.3; PyTorch 2.11.0+cu128; CUDA runtime 12.8 \\
Primary hardware & NVIDIA L40S GPU \\
Detailed traceability & Supplementary Material S1 \\
\bottomrule
\end{tabularx}
\end{table}

The frozen predictions and exports provide the basis for the manuscript's quantitative results. In the diagnostic export, the full \texttt{neuro\_symbolic} reference is taken from the main comparison and is not an independent rerun.

Supplementary Material S1 reports the recorded provenance identifiers, integrity records, environment freeze, artifact inventory, execution sequence, metric-regeneration checks, cross-suite reconciliation, and claim-to-evidence mapping. The reproducibility package links the benchmark generator, canonical source log, family and semantic specifications, accepted corpus, task-materialization code, frozen split, resolved configurations, saved predictions, result exports, and environment records. These artifacts support reconstructing the released benchmark and exactly regenerating reported metrics from archived predictions. They do not establish robustness to alternative benchmark generations, splits, or software environments.

\section{Extended Results}
\label{app:extended-results}

This appendix decomposes the aggregate macro-F1 point estimates in Table~\ref{tab:main-benchmark-results}. Values are averaged across the three saved seed evaluations. Early-warning target summaries are additionally averaged across leads, and lead summaries across targets.

\begin{table*}[p]
\centering
\caption{Early-warning macro-F1 by target, averaged across leads. Bold denotes the highest point estimate, not statistical significance.}
\label{tab:extended-ew-target}
\small
\begin{tabular}{@{}lccccc@{}}
\toprule
\textbf{Target} & \textbf{Raw MLP} & \textbf{Raw BiGRU} & \textbf{Raw Transformer} & \textbf{Contract-lite} & \textbf{Semantic BiGRU} \\
\midrule
$p_1$ & 0.5891 & 0.6050 & 0.6027 & 0.5488 & \textbf{0.7243} \\
$p_2$ & 0.6295 & 0.6412 & 0.6326 & 0.6635 & \textbf{0.6888} \\
$p_3$ & 0.6252 & 0.5991 & 0.6297 & 0.5601 & \textbf{0.6489} \\
$z$ & 0.6145 & 0.6154 & 0.5933 & 0.6066 & \textbf{0.6821} \\
\bottomrule
\end{tabular}
\end{table*}

\begin{table*}[p]
\centering
\caption{Early-warning macro-F1 by forecasting lead, averaged across targets.}
\label{tab:extended-ew-lead}
\small
\begin{tabular}{@{}lccccc@{}}
\toprule
\textbf{Lead} & \textbf{Raw MLP} & \textbf{Raw BiGRU} & \textbf{Raw Transformer} & \textbf{Contract-lite} & \textbf{Semantic BiGRU} \\
\midrule
3 & 0.6643 & 0.6599 & 0.6692 & 0.6695 & \textbf{0.7649} \\
6 & 0.6324 & 0.6439 & 0.6207 & 0.6044 & \textbf{0.6988} \\
12 & 0.5471 & 0.5417 & 0.5537 & 0.5104 & \textbf{0.5944} \\
\bottomrule
\end{tabular}
\end{table*}

\begin{table*}[p]
\centering
\caption{Pattern-detection macro-F1 by offline annotation target.}
\label{tab:extended-pd-target}
\small
\begin{tabular}{@{}lccccc@{}}
\toprule
\textbf{Target} & \textbf{Raw MLP} & \textbf{Raw BiGRU} & \textbf{Raw Transformer} & \textbf{Contract-lite} & \textbf{Semantic BiGRU} \\
\midrule
$p_1$ & 0.6103 & 0.6991 & 0.6571 & 0.5454 & \textbf{0.8446} \\
$p_2$ & 0.6873 & 0.7107 & 0.6813 & 0.7584 & \textbf{0.8462} \\
$p_3$ & 0.5594 & 0.5260 & 0.5788 & 0.5685 & \textbf{0.6478} \\
$z$ & 0.6638 & 0.6806 & 0.6555 & 0.6787 & \textbf{0.8010} \\
\bottomrule
\end{tabular}
\end{table*}

The semantic-input recurrent reference has the highest target- and lead-averaged point estimates in Tables~\ref{tab:extended-ew-target}--\ref{tab:extended-pd-target}. Table~\ref{tab:ew-complete-grid} reports the complete target--lead grid, while Table~\ref{tab:ew-test-support} reports the corresponding positive/total early-warning windows in the canonical test partition.
The complete grid nevertheless contains four settings led by comparator configurations, showing that the aggregate ordering is not uniform.

\begin{table*}[p]
\centering
\caption{Complete early-warning macro-F1 by target and lead. Bold denotes the row-wise largest point estimate, not significance.}
\label{tab:ew-complete-grid}
\small
\begin{tabular}{@{}lcccccc@{}}
\toprule
\textbf{Target} & \textbf{Lead} & \textbf{Raw MLP} & \textbf{Raw BiGRU} & \textbf{Raw Transformer} & \textbf{Contract-lite} & \textbf{Semantic BiGRU} \\
\midrule
$z$ & 3 & 0.6804 & 0.6753 & 0.6692 & 0.7000 & \textbf{0.7756} \\
$z$ & 6 & 0.6287 & 0.6308 & 0.6025 & 0.6221 & \textbf{0.7051} \\
$z$ & 12 & 0.5345 & 0.5401 & 0.5082 & 0.4978 & \textbf{0.5655} \\
$P_1$ & 3 & 0.6457 & 0.6648 & 0.6665 & 0.6334 & \textbf{0.7863} \\
$P_1$ & 6 & 0.6212 & 0.6489 & 0.6003 & 0.5200 & \textbf{0.7379} \\
$P_1$ & 12 & 0.5004 & 0.5013 & 0.5411 & 0.4930 & \textbf{0.6486} \\
$P_2$ & 3 & 0.7224 & 0.7362 & 0.7105 & 0.7831 & \textbf{0.8175} \\
$P_2$ & 6 & 0.6472 & 0.6686 & 0.6181 & \textbf{0.7189} & 0.7128 \\
$P_2$ & 12 & 0.5189 & 0.5188 & \textbf{0.5692} & 0.4887 & 0.5362 \\
$P_3$ & 3 & 0.6086 & 0.5633 & 0.6307 & 0.5615 & \textbf{0.6802} \\
$P_3$ & 6 & 0.6325 & 0.6274 & \textbf{0.6621} & 0.5566 & 0.6394 \\
$P_3$ & 12 & \textbf{0.6344} & 0.6066 & 0.5961 & 0.5621 & 0.6271 \\
\bottomrule
\end{tabular}
\end{table*}

\begin{table*}[p]
\centering
\caption{Positive/total early-warning windows in the canonical test partition. Balanced run labels do not imply balanced temporal targets.}
\label{tab:ew-test-support}
\small
\begin{tabular}{@{}lccc@{}}
\toprule
\textbf{Target} & \textbf{Lead 3} & \textbf{Lead 6} & \textbf{Lead 12} \\
\midrule
$z$ & 185/2580 & 185/2490 & 185/2310 \\
$p_1$ & 34/2580 & 34/2490 & 34/2310 \\
$p_2$ & 102/2580 & 102/2490 & 102/2310 \\
$p_3$ & 49/2580 & 49/2490 & 49/2310 \\
\bottomrule
\end{tabular}
\end{table*}

\section{Semantic-Input and Shortcut Diagnostics}
\label{app:ablation-diagnostics}

This appendix reports diagnostic comparisons used to qualify, rather than causally decompose, the main configuration-level results. Table~\ref{tab:ablation-deltas} reports the macro-F1 differences relative to the full semantic-input recurrent reference.

Zeroing either trajectory block lowers the aggregate point estimate within the evaluated semantic-input architecture. This establishes conditional signal utility, not an interaction effect. The shuffled operation is task-dependent: early-warning windows are permuted internally, complete runs are permuted for run classification, and pattern detection permutes all 96 rows before extracting windows at their original indices. The pattern-detection decrease therefore reflects changed order, content, and alignment, including originally future states; it does not isolate temporal order.

\begin{table*}[p]
\centering
\caption{Macro-F1 differences relative to the full semantic-input recurrent reference.}
\label{tab:ablation-deltas}
\small
\begin{tabularx}{\linewidth}{
  @{}
  >{\raggedright\arraybackslash}p{0.34\linewidth}
  *{3}{>{\centering\arraybackslash}X}
  @{}
}
\toprule
\textbf{Configuration} &
\textbf{Early warning} &
\textbf{Pattern detection} &
\textbf{Run classification} \\
\midrule
\texttt{capability\_only} & $-0.0227$ & $-0.0437$ & $-0.0699$ \\
\texttt{direct\_impact\_only} & $-0.0246$ & $-0.0491$ & $-0.1179$ \\
\texttt{shuffled\_semantic} & $-0.0106$ & $-0.2427$ & $-0.0555$ \\
\bottomrule
\end{tabularx}
\end{table*}

\begin{table*}[p]
\centering
\caption{Run-classification shortcut diagnostics. Diagnostic models are distinct from the main neural references.}
\label{tab:leakage-diagnostics}
\small
\begin{tabularx}{\linewidth}{
  @{}
  >{\raggedright\arraybackslash}p{0.36\linewidth}
  >{\raggedright\arraybackslash}X
  >{\centering\arraybackslash}p{0.17\linewidth}
  @{}
}
\toprule
\textbf{Diagnostic} & \textbf{Input view or feature group} & \textbf{Macro-F1} \\
\midrule
Full raw global-ID surface & Raw events with global identifiers & 0.8990 \\
Run-local alias surface & Primary aliased event view & 0.5623 \\
Sparse no-entity surface & Timestamp, event type, and subtype only & 0.3660 \\
Severity/confidence summary & Severity- and confidence-only scalars & 0.4949 \\
Target alias recurrence & Run-local target recurrence & 0.5662 \\
Location alias recurrence & Run-local location recurrence & 0.6997 \\
Combined alias recurrence & Target and location recurrence & 0.7321 \\
\bottomrule
\end{tabularx}
\end{table*}

The full raw surface is highly predictive in this run-classification diagnostic. Run-local aliasing removes stable global identities by construction, but recurrence summaries remain predictive. These diagnostic scores do not establish which features the main models use. The executed early-warning surface diagnostics are reported in Table~\ref{tab:ew-surface-diagnostics}.

\begin{table*}[p]
\centering
\caption{Executed early-warning surface diagnostics using standardized nearest-centroid classifiers.}
\label{tab:ew-surface-diagnostics}
\small
\begin{tabular}{@{}lccccc@{}}
\toprule
\textbf{Target} & \textbf{Lead} & \textbf{Global-ID} &
\textbf{Run-local aliases} & \textbf{Aliases + coarse} &
\textbf{Sparse no-entity} \\
\midrule
$z$ & 3 & 0.6696 & 0.5930 & 0.6323 & 0.6422 \\
$z$ & 6 & 0.6345 & 0.5699 & 0.6112 & 0.5966 \\
$z$ & 12 & 0.6084 & 0.5244 & 0.5463 & 0.5550 \\
$p_1$ & 3 & 0.6509 & 0.5213 & 0.5492 & 0.5426 \\
$p_1$ & 6 & 0.6257 & 0.5392 & 0.5615 & 0.5459 \\
$p_1$ & 12 & 0.5570 & 0.5484 & 0.5445 & 0.5443 \\
$p_2$ & 3 & 0.7236 & 0.6493 & 0.7012 & 0.6918 \\
$p_2$ & 6 & 0.6506 & 0.5944 & 0.6204 & 0.6220 \\
$p_2$ & 12 & 0.5821 & 0.4981 & 0.5643 & 0.5493 \\
$p_3$ & 3 & 0.6687 & 0.5959 & 0.5965 & 0.5915 \\
$p_3$ & 6 & 0.6739 & 0.6101 & 0.6242 & 0.6005 \\
$p_3$ & 12 & 0.6551 & 0.6108 & 0.6136 & 0.6255 \\
\bottomrule
\end{tabular}
\end{table*}

The recurrence rows in Table~\ref{tab:leakage-diagnostics} apply only to run classification. No pattern-detection surface diagnostic or temporal-task recurrence diagnostic was executed. The sparse no-entity view also removes actor, confidence, and severity, so it is not an entity-only intervention.

\FloatBarrier
\section{Grouped-Bootstrap and Setting-Level Analyses}
\label{app:bootstrap-analyses}

The pipeline applies 1,000 grouped resamples to fixed saved predictions, using \texttt{run\_id} as the resampling unit. Lower and upper endpoints are computed separately for each seed and task setting and then averaged. Thus, the reported bounds are descriptive summaries; they are not confidence intervals from a joint bootstrap of the headline task aggregates or the three fitted models. Table~\ref{tab:bootstrap-headline} reports the resulting mean setting-level point estimates and grouped-bootstrap uncertainty bounds.

\begin{table}[H]
\centering
\caption{Mean setting-level macro-F1 point estimates and grouped-bootstrap 95\% uncertainty bounds.}
\label{tab:bootstrap-headline}
\small
\begin{tabular}{@{}lccc@{}}
\toprule
\textbf{Configuration} & \textbf{Early warning} & \textbf{Pattern detection} & \textbf{Run classification} \\
\midrule
\texttt{simple\_neural} & 0.6146 [0.5259, 0.6779] & 0.6302 [0.5444, 0.7127] & 0.7550 [0.5909, 0.8883] \\
\texttt{advanced\_neural} & 0.6152 [0.5247, 0.6847] & 0.6541 [0.5629, 0.7360] & 0.7424 [0.5678, 0.8878] \\
\texttt{transformer\_raw\_event} & 0.6145 [0.5218, 0.6870] & 0.6432 [0.5538, 0.7293] & 0.6931 [0.5160, 0.8481] \\
\texttt{semantic\_baseline} & 0.5948 [0.5432, 0.6371] & 0.6378 [0.5567, 0.7210] & 0.5928 [0.3973, 0.7664] \\
\texttt{neuro\_symbolic} & \textbf{0.6860} [0.5875, 0.7650] & \textbf{0.7849} [0.6730, 0.8558] & \textbf{0.7865} [0.6196, 0.9218] \\
\bottomrule
\end{tabular}
\end{table}

The semantic-input recurrent reference has the highest headline point estimate on all three tasks, but the uncertainty summaries overlap broadly. Because the procedure does not jointly resample the headline aggregates and omits split, retraining, and benchmark-generation variability, these bounds do not support statistically resolved superiority.

The shuffled pattern-detection condition has point estimate 0.5422 with mean bounds [0.4919, 0.5928], compared with 0.7849 [0.6730, 0.8558] for the full semantic-input recurrent reference. This comparison characterizes permutation of the complete trajectory before window extraction. Because the operation changes window content and alignment, it cannot isolate a temporal-order effect.

\begin{table}[H]
\centering
\caption{Early-warning settings in which the semantic-input recurrent reference does not have the highest point estimate.}
\label{tab:bootstrap-setting-exceptions}
\small
\begin{tabular}{@{}lclcc@{}}
\toprule
\textbf{Target} & \textbf{Lead} & \textbf{Leading configuration} & \textbf{Leader} & \textbf{Semantic BiGRU} \\
\midrule
$p_2$ & 12 & \texttt{transformer\_raw\_event} & 0.5692 & 0.5362 \\
$p_2$ & 6 & \texttt{semantic\_baseline} & 0.7189 & 0.7128 \\
$p_3$ & 12 & \texttt{simple\_neural} & 0.6344 & 0.6271 \\
$p_3$ & 6 & \texttt{transformer\_raw\_event} & 0.6621 & 0.6394 \\
\bottomrule
\end{tabular}
\end{table}

Table~\ref{tab:bootstrap-setting-exceptions} lists the early-warning settings in which the semantic-input recurrent reference does not have the highest point estimate.

These exceptions demonstrate setting-level heterogeneity without changing the aggregate point-estimate ordering.

\clearpage

\bibliographystyle{unsrtnat}
\bibliography{bibliography}

\end{document}